\documentclass[
twocolumn,
]{ceurart}

\usepackage{listings}
\begin{document}

\copyrightyear{2022}
\copyrightclause{Copyright for this paper by its authors.
  Use permitted under Creative Commons License Attribution 4.0
  International (CC BY 4.0).}

\conference{LLM4AI'23: Workshop on Foundations and Applications in Large-scale AI Models
-Pre-training, Fine-tuning, and Prompt-based Learning, co-located with the 29TH ACM SIGKDD CONFERENCE ON KNOWLEDGE DISCOVERY AND DATA MINING (KDD), August 6-10, 2023, Long Beach, CA, USA}

\title{Improving Small Language Models on PubMedQA via Generative Data Augmentation}

\author[1]{Zhen Guo}[%
%orcid=0000-0002-1347-3451,
email=zguo0525@mit.edu,
%url=https://zguo0525.github.io/,
]
\cormark[1]
\address[1]{MIT Department of Electrical Engineering and Computer Science,
  77 Massachusetts Ave, Cambridge, MA 02139}
% \address[2]{MIT Computer Science \& Artificial Intelligence Laboratory,
%   77 Massachusetts Ave, Cambridge, MA 02139}

\author[1]{Yanwei Wang}[%
%orcid=0000-0001-7116-9338,
email=yanwei@mit.edu,
% url=https://kmitd.github.io/ilaria/,
]

\author[1]{Peiqi Wang}[%
%orcid=0000-0002-9421-8566,
email=wpq@mit.edu,
% url=http://conceptbase.sourceforge.net/mjf/,
]

\author[1]{Shangdi Yu}[%
%orcid=0000-0002-8907-692X,
email=shangdiy@mit.edu,
% url=http://conceptbase.sourceforge.net/mjf/,
]

\cortext[1]{Corresponding author.}

\begin{abstract}
Large Language Models (LLMs) have made remarkable advancements in the field of natural language processing. However, their increasing size poses challenges in terms of computational cost. On the other hand, Small Language Models (SLMs) are known for their efficiency, but they often struggle with limited capacity and training data, especially in specific domains. In this paper, we introduce a novel method aimed at improving SLMs in the medical domain using LLM-based generative data augmentation. The objective of our approach is to develop more efficient and capable models that are specifically tailored for specialized applications. Through experiments conducted on the PubMedQA dataset, we demonstrate the effectiveness of LLMs in refining and diversifying existing question-answer pairs. This refinement process leads to improved performance in a significantly smaller model after fine-tuning. Notably, our best SLM, with under 1.6 billion parameters, outperforms the few-shot GPT-4 on the PubMedQA dataset. Our code and generated data are publicly available to facilitate further explorations~\cite{2023drllama}.
\end{abstract}

\begin{keywords}
large language models \sep 
small language models \sep
medical question-answering \sep
data augmentation
\end{keywords}

\maketitle

\section{Introduction}
In recent years, Large Language Models (LLMs) have transformed the field of natural language processing, demonstrating exceptional performance in a wide range of tasks. These models, powered by massive amounts of data and extensive pre-training~\cite{hoffmann2022training}, have advanced the state-of-the-art in various applications such as machine translation, program synthesis, and question-answering~\cite{zhao2023survey, austin2021program, chen2021evaluating}. Although LLMs have impressive capabilities, their growing size presents challenges in terms of computational efficiency, particularly for real-world applications and domain-specific tasks~\cite{strubell2019energy, gu2021domain}. Problems such as medical question-answering or legal document analysis require specialized knowledge that may not be fully captured by general-purpose LLMs~\cite{ram2021few, sun2023short}.

Small Language Models (SLMs), on the other hand, offer a more computationally efficient alternative to LLMs. However, SLMs often struggle in domain-specific tasks due to their limited capacity and training data. This limitation requires the development of new strategies to enhance the performance of SLMs in specialized tasks while maintaining their computational efficiency~\cite{poerner2020inexpensive, iandola2020squeezebert}.

In this paper, we introduce a novel method that improves SLMs in the medical domain through LLM-based generative data augmentation. The objective is to develop more efficient and capable models tailored for specialized medical applications without using billions of parameters. Our results in the PubMedQA dataset~\cite{jin2019pubmedqa} demonstrate the effectiveness of LLM in the refinement and diversification of question-answer pairs, leading to improved performance of a significantly smaller model after fine-tuning. The best SLM, with less than 1.6 billion parameters, outperforms the few-shot GPT-4 on PubMedQA, as shown in Table~\ref{tab:zero}. In general, our method holds promise for enhancing SLMs for medical tasks, bridging the gap between computational efficiency and model performance in specialized domains.

\begin{table}[h!]
    \vspace{-2pt}
  \caption{ChatGPT vs. BioGPT with fine-tuning on PubMedQA} %\shangdi{should we remove "Zero-shot" in the caption? Since we have a 5-shot GPT-4 in the table. And the number of GPT3.5 here seems to be different from Table3?}}
  \label{tab:zero}
    \begin{tabular}{lcc}
    \toprule
    \textbf{Model} & \textbf{Accuracy} & \textbf{Macro-F1}\\
    \midrule
    GPT-3.5-turbo (175B) & 0.372 & 0.327 \\
    GPT-4 (0-shot)~\cite{nori2023capabilities} & 0.752 & NA \\
    GPT-4 (5-shot)~\cite{nori2023capabilities} & 0.744 & NA \\
    Best BioGPT (1.6B) & 0.754 & 0.520\\
    Human Performance~\cite{jin2019pubmedqa} & 0.780 & 0.722 \\
    \bottomrule
\end{tabular}
\vspace{-6pt}
\end{table}

\section{Technical background}
\subsection{Efficient fine-tuning}

Fine-tuning LLMs for specific tasks poses computational and time-related challenges~\cite{liu2022few, vos2022towards}. To address these issues, researchers have developed efficient fine-tuning techniques, such as Prefix Tuning and Low-rank Adaptation, as alternatives to traditional fine-tuning methods that update the model's weights entirely. Prefix tuning~\cite{li2021prefix} adapts the behavior of a language model to specific tasks without modifying its pre-trained weights. While the low-rank adaptation~\cite{hu2021lora} allows the model to capture the essential characteristics of the data and adapt to domain-specific tasks effectively by decomposing the weight matrices into smaller matrices.

\subsection{Data Augmentation using LLMs}
LLMs serve as powerful tools to generate realistic text samples based on existing data. For NLP tasks, generating data with LLM can involve paraphrasing text, creating alternative question-answer pairs, or generating new sentences~\cite{edwards2021guiding}. Producing diverse representations of input data enables models to learn various ways to express the same underlying concepts, increasing their adaptability to real-world data variations.

% \begin{figure}[ht!]
%   \centering
%   % \includegraphics[width=0.4\textwidth]{rewrite.png}
%   \includegraphics[width=\columnwidth]{qa.pdf}
%   \caption{Zero-shot prompting to generative new data on the PubMedQA dataset with OpenAI API}
%   \label{fig:Generative}
% \end{figure}

For our preliminary study on the PubMedQA dataset, we used GPT-3.5 Turbo and GPT-4 to either rewrite existing medical question-answering pairs or generate new pairs from the training dataset (with a size of 450) with zero-shot prompting. This approach helped improve the diversity and coverage of the training data, ultimately improving the performance of the medical question-answering model trained on the augmented dataset.

\section{Experimental settings} 
We performed experiments on the MIT Supercloud~\cite{reuther2018interactive}, using PyTorch 1.12 and Python 3.8 with eight NVIDIA V100 GPUs and Intel Xeon Gold 6248 processors. We investigated the effectiveness of prefix tuning and low-rank adaptation on BioGPT-Large~\cite{luo2022biogpt}, LLaMA-7b~\cite{touvron2023llama}, and Alpaca-7b~\cite{alpaca} for medical question-answering tasks. The evaluation was carried out on the PubMedQA dataset~\cite{jin2019pubmedqa}, splitting it into 450 training, 50 validation, and 500 test samples. Accuracy and F1 score were calculated based on a hard match between predicted and ground truth answers. For prefix tuning, we follow the original implementation~\cite{li2021prefix} and explored a token range of 16 to 512, while low-rank adaptation varied alpha from 16 to 512 with a fixed rank of 4. Fine-tuning employed a learning rate of 5e-5, AdamW optimizer~\cite{loshchilov2017decoupled}, linear warm-up scheduler~\cite{vaswani2017attention}, gradient accumulation of 32 steps~\cite{lin2017deep}, and a batch size of 1024 tokens. During inference, we applied techniques including Repetition Penalty Logits Processor (penalty factor of 2.0), Temperature Logits Warper (temperature of 0.8), and beam search decoding with a beam size of 5 to ensure output quality. 

\section{Results}
\subsection{Low-rank Adaptation outperforms Prefix Tuning}
\vspace{-12pt}
\begin{figure}[ht!]
  \centering
    \includegraphics[width=0.77\columnwidth]{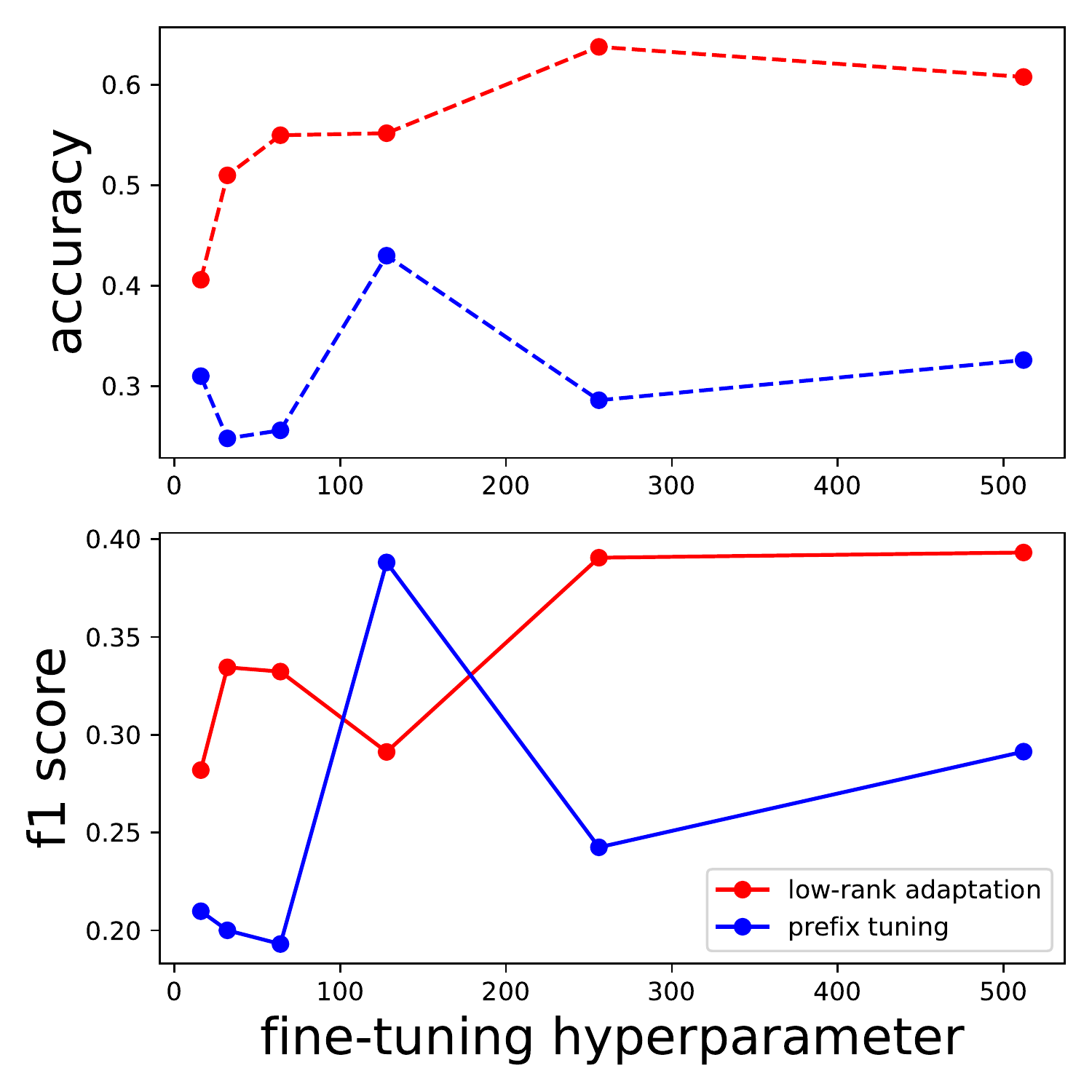}
    \vspace{-8pt}
  \caption{Comparison between two fine-tuning techniques for BioGPT-Large.}
  \label{fig:line_plot}
  \vspace{-4pt}
\end{figure}

We compared the performance of two techniques, Low-rank Adaptation and Prefix Tuning, for BioGPT-Large (Figure~\ref{fig:line_plot}). We observed that Low-rank Adaptation demonstrated stability across with different hyperparameters (16 to 512), while Prefix Tuning showed sensitivity to the virtual token range. This finding suggests that Low-rank adaptation is more robust and less sensitive to hyperparameter selection, providing consistent and reliable performance for efficient fine-tuning. For all the results below, Low-rank adaptation is the default fine-tuning technique.

\subsection{Instruction-tuning constrains domain adaptability of language models}

In Table~\ref{baseline}, we present a comparison of BioGPT-Large, LLaMA-7b, and Alpaca-7b, all fine-tuned on the original PubMedQA dataset without data augmentation as the baseline. Alpaca-7b, a derivative of LLaMA-7b, is an instruction-tuned LLM designed to improve task-specific performance by following instructions. However, this approach restricts its adaptability to other domain-specific tasks compared to a naïve pre-trained model. In our experiments, LLaMA-7b shows superior generalizability compared to BioGPT-Large by exhibiting a higher F1 score, when fine-tuned only on the original PubMedQA dataset. The reported accuracy for BioGPT-Large is lower than the numbers reported by \citet{luo2022biogpt} because we have different fine-tuning settings. While \citet{luo2022biogpt} inserted the virtual tokens right before the answer token, we inserted the virtual tokens before the question token to avoid the risk of overfitting.
 
\begin{table}[ht!]
\centering
\caption{Comparison of BioGPT-Large, LLaMA-7b and Alpaca-7b fine-tuned on the original PubMedQA dataset.}
\begin{tabular}{lcc}
\toprule
\textbf{Model} & \textbf{Accuracy} & \textbf{Macro-F1}\\
\midrule
BioGPT-Large & 0.630 & 0.387 \\
LLaMA-7b & 0.594 & 0.495 \\
Alpaca-7b & 0.380 & 0.335\\
\bottomrule
\end{tabular}
\label{baseline}
\vspace{-8pt}
\end{table}

\subsection{Comparison between generative data augmentation approaches}

\begin{table}[t]
\centering
\caption{Comparison of LLaMA-7b and BioGPT-Large fine-tuned on augmented PubMedQA (\normalfont \textit{Acc.} stands for accuracy, \textit{GPT-3.5} refers to GPT-3.5-turbo, \textit{BioGPT} represents BioGPT-Large, and \textit{F1} denotes marco F1 Score).}
\footnotesize
\begin{tabular}{p{0.7cm}llll}
\toprule
\textbf{SLM} & \textbf{LLM} & \textbf{Augment.} & \textbf{Acc.} (best) & \textbf{F1} (best)\\
\midrule
      \multirow{7}{*}{LLaMA} &  \multirow{4}{*}{GPT-3.5} & \text{none} & 0.594 & 0.495\\
      & & \textbf{rewriteQA} & \textbf{0.642} & \textbf{0.497}\\
      & & \text{newQA} & 0.552 & 0.460\\
      & & \text{combinedQA} & 0.582 & 0.485\\
      % & \text{more-combinedQA} & ? & ?\\
      \cline{2-5}
        & \multirow{3}{*}{GPT-4} & \text{rewriteQA} & 0.540 & 0.463\\
      & & \textbf{newQA} & \textbf{0.576} & \textbf{0.451}\\
      & & \text{combinedQA} & 0.506 & 0.446\\
      
      \midrule
      \multirow{7}{*}{\text{BioGPT}} & \multirow{4}{*}{GPT-3.5} & \text{none} & 0.630 & 0.387\\
      & & \textbf{rewriteQA} & \textbf{0.720} & \textbf{0.498}\\
      & & \text{newQA} & 0.718 & 0.491\\
      & & \text{combinedQA} & 0.714 & 0.493\\
      % & \text{more-combinedQA} & 0.576 & 0.487\\
      \cline{2-5}
        & \multirow{3}{*}{GPT-4} & \text{rewriteQA} & 0.654 & 0.471\\
      & & \textbf{newQA} & \textbf{0.754} & \textbf{0.520}\\
      & & \text{combinedQA} & 0.708 & 0.518\\
      \bottomrule
      \label{full-compare}
\end{tabular}
\vspace{-20pt}
\end{table}

In Table~\ref{full-compare}, we provide a comparison of LLaMA-7b and BioGPT-Large fine-tuned on the augmented PubMedQA dataset. Our experiments demonstrate the efficacy of utilizing LLMs such as ChatGPT for refining and expanding question-answer pairs to enhance domain-specific QA datasets, even when the LLM exhibits near-random performance in generating answers (as for the case for gpt-3.5-turbo). The resulting alternative representations of questions and answers facilitated the construction of more diverse and robust training datasets suitable for SLMs.

However, we found that instructing an LLM (gpt-3.5-turbo) lacking domain knowledge to generate entirely new question-answer pairs did not lead to an improvement and resulted in a degradation of the downstream task performance for the fine-tuned SLM. This observation suggests that while LLMs are effective in refining and diversifying existing question-answer pairs, their ability to create novel, high-quality pairs for domain-specific tasks remains limited.

On the other hand, recent advances in LLMs such as GPT-4, which have domain-specific knowledge and question-answering capacity for PubMedQA, can generate useful new training data. By incorporating new question-answer pairs from GPT-4 into the training process, we can significantly improve the performance of the fine-tuned smaller models. This finding highlight the importance of LLMs with domain-specific knowledge in enhancing domain-specific QA datasets and improving the performance of downstream tasks.

Finally, not surprisingly, when BioGPT is fine-tuned on an augmented dataset, it outperforms LLaMA-7B. This is consistent with the previous finding~\cite{luo2022biogpt}, and highlights the effectiveness of pretraining with domain-specific data, enabling BioGPT to better understand and excel in domain-specific tasks. Leveraging domain-specific knowledge during fine-tuning improves the model's accuracy and contextual relevance, resulting in superior performance for domain-specific questions or tasks.

\section{Future works}
A promising direction for future work is to investigate the application of knowledge distillation, a popular technique that trains a smaller language model to mimic the behavior of a larger language model on medical question-answering tasks. 

Another potential approach is through contrastive learning. By training an SLM using contrastive learning on medical question-answering data, contrastive loss can help the model learn to identify similarities and differences between different instances of data and improve its ability to generalize to new and unseen data.

% In summary, future work should investigate the effectiveness of these approaches and explore their potential limitations in different contexts. The combination of generative data augmentation, knowledge distillation, and contrastive learning can significantly enhance the performance of SLMs on medical QA tasks and other domain-specific applications.

\section{Conclusion}
Our research highlights the effectiveness of LLM-based generative data augmentation in enhancing domain-specific question answering datasets. However, instructing LLMs without domain knowledge, such as GPT-3.5-turbo, to generate new question-answer pairs resulted in decreased performance for fine-tuned smaller models. Conversely, leveraging LLMs with domain-specific knowledge, like GPT-4, significantly improved the performance of fine-tuned models by generating valuable new training data. These findings underscore the importance of incorporating domain-specific knowledge when applying generative data augmentation techniques.

\begin{acknowledgments}
 We thank Prof. Yoon Kim at MIT CSAIL for his guidance and feedback. The authors acknowledge the MIT SuperCloud and Lincoln Laboratory Supercomputing Center for providing computing resources. We would also like to acknowledge OpenAI for providing access to their API. 
\end{acknowledgments}

%%
%% Define the bibliography file to be used
\bibliography{sample-ceur}

\end{document}